\documentclass[letterpaper]{article} 
\usepackage{aaai25}  
\usepackage{times}  
\usepackage{helvet}  
\usepackage{courier}  
\usepackage[hyphens]{url}  
\usepackage{graphicx} 
\usepackage{natbib}  
\usepackage{caption} 
\usepackage{algorithm}
\usepackage{algorithmic}
\usepackage{verbatim}
\usepackage{amssymb}
\usepackage{multirow}
\usepackage{subfigure}
\usepackage{amsmath}
\usepackage{mathrsfs}
\usepackage{graphicx}
\usepackage{multirow}
\usepackage{booktabs}
\usepackage{xcolor}

\usepackage{newfloat}
\usepackage{listings}
\DeclareCaptionStyle{ruled}{labelfont=normalfont,labelsep=colon,strut=off} 
\floatstyle{ruled}
\newfloat{listing}{tb}{lst}{}
\floatname{listing}{Listing}
\title{Boosting Short Text Classification with Multi-Source Information Exploration and Dual-Level Contrastive Learning}
\author{
    Yonghao Liu\textsuperscript{\rm 1}\equalcontrib, Mengyu Li\textsuperscript{\rm 1}\equalcontrib, Wei Pang\textsuperscript{\rm 2}, Fausto Giunchiglia\textsuperscript{\rm 3}, \\ 
    Lan Huang\textsuperscript{\rm 1}, Xiaoyue Feng\textsuperscript{\rm 1}\thanks{Corresponding Author.}, Renchu Guan\textsuperscript{\rm 1}\footnotemark[2]
}
\affiliations{
    \textsuperscript{\rm 1}Key Laboratory of Symbolic Computation and Knowledge Engineering of the Ministry of \\ Education, College of Computer Science and Technology, Jilin University \\
    \textsuperscript{\rm 2}Mathematical and Computer Sciences, Heriot-Watt University \\
    \textsuperscript{\rm 3}University of Trento \\

    \{yonghao20, mengyul21\}@mails.jlu.edu.cn,
w.pang@hw.ac.uk, fausto.giunchiglia@unitn.it \\
\{huanglan, fengxy, guanrenchu\}@jlu.edu.cn
}

\usepackage{bibentry}

\begin{document}

\maketitle

\begin{abstract}
Short text classification, as a research subtopic in natural language processing, is more challenging due to its semantic sparsity and insufficient labeled samples in practical scenarios. We propose a novel model named MI-DELIGHT for short text classification in this work. Specifically, it first performs multi-source information (\textit{i.e., statistical information, linguistic information, and factual information}) exploration to alleviate the sparsity issues. Then, the graph learning approach is adopted to learn the representation of short texts, which are presented in graph forms. Moreover, 
we introduce a dual-level (\textit{i.e., instance-level and cluster-level}) contrastive learning auxiliary task to effectively capture different-grained contrastive information within massive unlabeled data. Meanwhile, previous models merely perform the main task and auxiliary tasks in parallel, without considering the relationship among tasks. Therefore, we introduce a hierarchical architecture to explicitly model the correlations between tasks. We conduct extensive experiments across various benchmark datasets, demonstrating that MI-DELIGHT significantly surpasses previous competitive models. It even outperforms popular large language models on several datasets. 
\end{abstract}

%
\section{Introduction}
Text classification is a fundamental task in natural language processing (NLP). As a special form of text, short texts often appear in our daily life in the form of tweets, queries, and news feeds \cite{phan2008learning}. To deal with these short texts, short text classification (STC), as a subtask of text classification, has attracted extensive attention from the research community. It is widely used in various practical scenarios, such as news classification 
\cite{dilrukshi2013twitter}, sentiment analysis \cite{chen2019deep}, and query intent classification \cite{wang2017combining}. It is worth noting that compared to traditional text classification, STC is particularly nontrivial, which is mainly attributed to its two well-known challenges, \textit{i.e.}, \textit{semantic sparsity} and \textit{limited labeled texts} \cite{linmei2019heterogeneous}.

For the challenge of \textit{semantic sparsity}, short texts typically contain only one or two sentences with few words, which have limited available contextual information \cite{tang2015pte}. Such severe semantic sparsity often leads to vagueness and ambiguity, thus hindering the accurate understanding of short texts. An effective solution is to explore multi-source information to enrich the context for short texts. On the one hand, we can collect the \textit{statistical information} and \textit{linguistic information} contained within short texts. \textit{Statistical information} is related to the statistics of words that constitute texts, and is often obtained by modeling word co-occurrence and word distribution probabilities in the text \cite{thilakaratne2019systematic, liu2024resolving}. For example, by analyzing the word statistics in the text, Latent Dirichlet Allocation (LDA) can uncover the topic structure that can enrich the information in short texts. 
\textit{Linguistic information} is implicit in the syntax and semantics of texts \cite{liu2019linguistic}. For instance, we can obtain the part-of-speech (POS) information of words to determine their syntactic roles in the text. 
On the other hand, auxiliary \textit{factual information} can also be injected to compensate for the missing contextual information \cite{liu2023time, liu2023global}. In this paper, factual information mainly refers to those text-related entities existing in common sense knowledge graphs \cite{chen2019deep}. With such enriched auxiliary information, the learned model can naturally understand the meanings of short texts better. 

When faced with the challenge of limited labeled texts in real-world applications, it is often the case that there is a vast amount of easily accessible short texts, but only a small number of labeled data are available \cite{kenter2015short}. In addition, the proportion of unlabeled data is much higher compared to that of long texts. Consequently, deep learning models that rely on large-scale labeled data for training are susceptible to overfitting issues, leading to unsatisfactory performance outcomes. To cope with such issue, \textit{on the one hand}, some works \cite{linmei2019heterogeneous,yang2021hgat} are mainly devoted to fully utilizing limited labeled short texts. They perform supervised graph learning on the constructed corpus-level graph to learn the textual representations. Nevertheless, the performance of these models is largely influenced by the limited labeled data, as they only provide relatively restricted information. 
\textit{On the other hand}, some works \cite{liu2016recurrent,liu2017adversarial} attempt to introduce auxiliary tasks to alleviate the inefficient data problem. They typically design auxiliary tasks and then jointly train these tasks, aiming to enable the knowledge contained in the tasks to be utilized by other tasks, thereby improving the model generalization ability. However, the reliability of auxiliary tasks are questionable, and unreliable auxiliary ones can impair the model performance. 

Recently, contrastive learning (CL) has attracted tremendous attention due to its effectiveness in extracting features from unlabeled data and simple mechanism. Using CL for auxiliary feature learning appears to be a promising approach, as it enables the extraction of self-supervised contrastive information from a large corpus of unlabeled texts. Moreover, CL has been extensively demonstrated to function as a dependable auxiliary task for extracting discriminative information \cite{chen2022contrastnet,pan2022improved}. By this way, we can simultaneously handle the limitations of the two aforementioned types of models. 
However, typically, only instance-level contrastive learning (ICL) is previously used for auxiliary feature learning, which regards each instance as a distinct class. The unique positive pair originates from the same instance, and other instances sharing similar underlying semantics are considered negative pairs and pushed apart. 
Therefore, it is not sufficient to use such an unsupervised ICL approach from a fine-grained perspective alone. We also need to introduce a coarse-grained CL as an auxiliary task, such as cluster-level supervised CL (CCL), which can further enable the aggregation of samples that share intrinsically similar signals from a coarse-grained perspective. 
Furthermore, previous models that incorporate CL simply combine the losses of the main task and the auxiliary task, performing them in parallel, without adequately considering the significance of a well-structured architecture that facilitates connections among multiple tasks. This approach is deemed unreasonable, as there exists a causal relationship established through the learned features across tasks. In other words, as we progress from ICL to CCL to classification, the growing complexity of the tasks necessitates the acquisition of increasingly sophisticated features, enabling a transition from rudimentary to abstract characteristics.

In this paper, we introduce a novel model called \textbf{MI-DELIGHT} that leverages \textbf{M}ulti-source \textbf{I}nformation and \textbf{D}ual-level contrastiv\textbf{E} \textbf{L}earn\textbf{I}ng for s\textbf{H}ort \textbf{T}ext classification. On the one hand, graphs, as a basic data structure, possess the desirable characteristics of flexibility and simplicity. As such, we adopt graphs as the uniform representation form for texts with injected information. On the other hand, graph neural networks (GNNs) have a natural advantage in learning from graph data. Meanwhile, in numerous NLP tasks, GNNs have demonstrated superior performance in capturing non-consecutive and long-range word interactions, as well as their powerful representation capabilities for modeling texts. Therefore, we first construct a word graph and a POS graph to explore the statistical and linguistic information contained in short texts. Additionally, we also build an entity graph to introduce supplementary factual information. After obtaining all the information mentioned above by GNNs and extracting rudimentary text features, we design a dual-level CL auxiliary task to assist in obtaining improved text features in a more comprehensive manner. Importantly, we introduce a hierarchical structure to leverage the casual relationships among tasks and extract abstract features step by step. Specifically, we first employ ICL based on the elementary text features to capture the fine-grained contrastive information. 
Then, we perform CCL based on advanced text features obtained during the ICL process 
to capture the coarse-grained contrastive information. Finally, we classify high-level text features obtained during the CCL process. 
In summary, our contributions are as follows:

(1) We propose a novel model, namely MI-DELIGHT, which is capable of modeling short texts and resolving existing semantic sparsity and inefficient labeled samples.

(2) We build three types of graphs to explore statistical, linguistic and factual information to compensate for critical context. Moreover, we design a hierarchical dual-level CL auxiliary tasks, including CCL and ICL, to effectively capture multi-grained contrastive information. 

(3) We conduct diverse experiments, and MI-DELIGHT consistently surpasses other competitive models, including some popular large language models, across several benchmark datasets.

\section{Related Work}
\noindent \textbf{Text Classification}: 
Traditional text classification methods typically first use hand-crafted lexical features \cite{li2022survey}, such as BoW and TF-IDF, to represent text and then adopt SVM or Naive Bayes classifiers. With the development of neural networks, deep learning models without feature engineering, such as convolutional neural networks (CNNs) and recurrent neural networks (RNNs), have become mainstream approaches in this area. Moreover, recent studies \cite{liu2021vpalg,liu2022few} have demonstrated the successful application of GNNs to text classification tasks. These models capture word interactions using graph structures and have shown promising results. One line of work involves building corpus graphs, treating both text and words as nodes, and performing classification in a semi-supervised manner \cite{yao2019graph,liu2020tensor}. Another line of research focuses on constructing a graph for each text and deriving document representations through graph learning on word-word edges \cite{ding2020more,liu2021deep}.

\noindent \textbf{Short Text Classification}:
STC is a challenging task in which irregular word orders and missing function words hinder the proper understanding of short texts. Some existing and popular approaches attempt to introduce additional information, such as entities or latent topics, to assist with text understanding \cite{zeng2018topic}. Some studies \cite{ye2020document,wang2021hierarchical} have conducted label propagation via graph structures of constructed heterogeneous graphs and yielded notable gains. Further, several models \cite{go2022contrastive, liu2024improved} propose leveraging CL on the corpus-level graph for STC and achieves promising results. However, these models only engage in instance-level CL, thus disregarding cluster-level features. Recent popular large language models (LLMs), such as GPT-3.5 \cite{ouyang2022training} and Llama \cite{touvron2023llama} have been pre-trained on massive high-quality data, thus exhibiting excellent understanding of general texts. However, their performance on domain-specific (\textit{e.g}, medical or legal domains) texts is not as expected \cite{chang2023survey}.

\noindent \textbf{Contrastive Learning}: 
CL approaches learn representations by contrasting positive pairs against negative pairs, and have been highly successful in various fields such as NLP \cite{gao2021simcse,wu2021esimcse} and graphs \cite{liu2024meta, liu2024simple}. Initially, many CL approaches focus on instance discrimination tasks in an unsupervised manner \cite{caron2020unsupervised,tian2020contrastive}. The following studies \cite{khosla2020supervised,liu2024meta,li2024simple} explore fully supervised CL, which can explicitly leverage label information, enabling the extraction of more task-relevant information. Some recent studies \cite{zheng2021weakly,huynh2022boosting} consider similar samples as positive pairs and aim to pull them together. However, these methods mostly focus on unsupervised tasks and do not take advantage of the handful of available instance labels. 

\section{Preliminary}
Most modern GNN models follow a recursive neighborhood aggregation scheme, where the representation of a node is iteratively updated by aggregating the features of its neighbors. A classic model is the graph convolutional network (GCN) \cite{kipf2016semi}, which can be defined formally as follows: 
\begin{equation}
\label{gnn}
    \mathbf{H}^{(\ell+1)} = \sigma(\hat{\mathbf{D}}^{-\frac{1}{2}} \hat{\mathbf{A}}\hat{\mathbf{D}}^{-\frac{1}{2}}\mathbf{H}^{(\ell)}\mathbf{W}^{(\ell)}),
\end{equation}
where $\mathbf{H}^{(\ell)}$ is the $\ell$-th output node representation and $\mathbf{H}^{(0)}=\mathbf{X}$ is an initial node embedding. $\hat{\mathbf{A}}=\mathbf{A}+\mathbf{I}$ is an adjacency matrix with added self-loops, and $\hat{\mathbf{D}}_{ii}=\sum_j\hat{\mathbf{A}}_{ij}$ is the corresponding diagonal degree matrix. $\sigma(\cdot)$ is an activation function such as ReLU and $\mathbf{W}^{(\ell)}$ is a layer-specific trainable matrix.

\section{Method}
\begin{figure*}
    \centering
    \includegraphics[width=\linewidth]{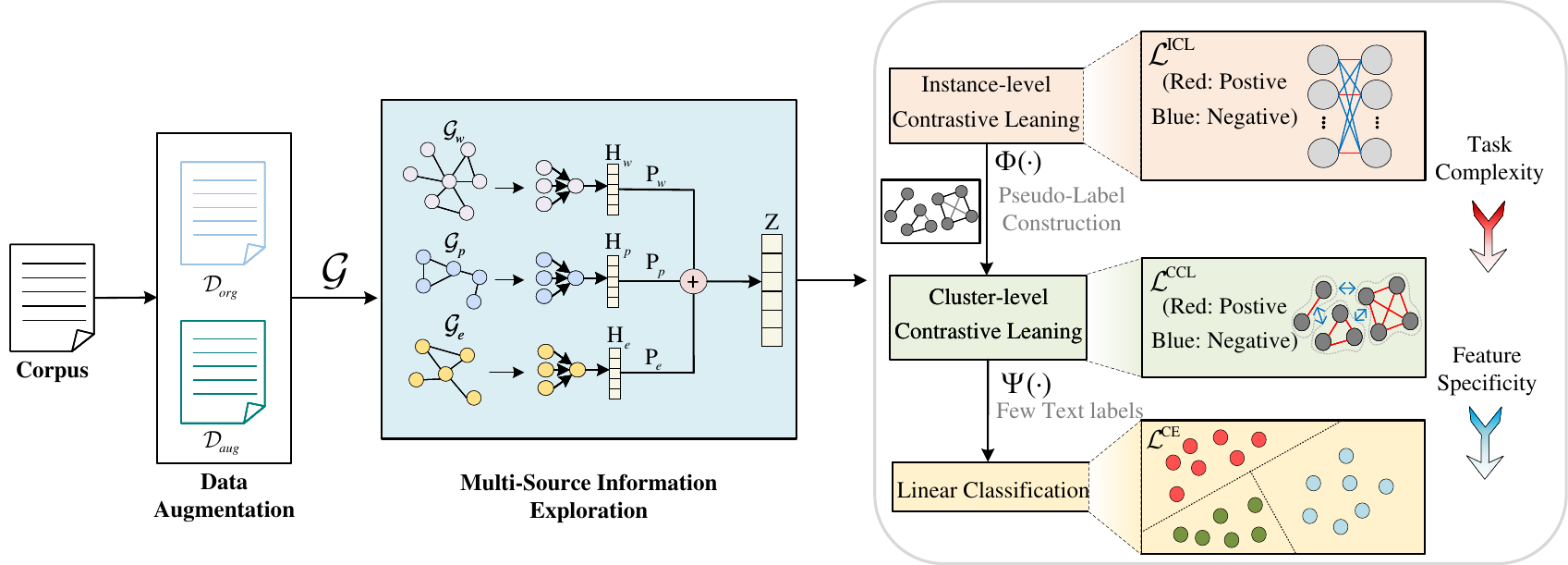}
    \caption{The overall architecture of MI-DELIGHT. We first generate augmented samples for the input texts. Then, the original corpus $\displaystyle \mathcal{D}_{\text{org}} =\left\{d_{i}^{\text{org}}\right\}_{i=1}^{N}  $ and the augmented corpus $\displaystyle \mathcal{D}_{\text{aug}} =\left\{d{_{i}^{\text{aug}}}\right\}_{i=1}^{N}$ are used to construct a word graph $\mathcal{G}_w$, a POS graph $\mathcal{G}_p$ and an entity graph $\mathcal{G}_e$, and the text embeddings $\mathbf{Z}$ are obtained via the text representation learning module. Finally, these embeddings are mapped through different projection heads into different hidden spaces to which ICL, CCL, and cross-entropy (CE) are applied in a certain hierarchical order. From ICL to CCL and then to CE, the task complexity keeps increasing, and the features keep more abstract. Here, \textit{feature specificity} represents the abstraction level of features.}
    \label{flowchart}
    \vspace{-0.5em}
\end{figure*}

In this section, we present the proposed MI-DELIGHT model for STC. The overall architecture is shown in Fig. \ref{flowchart}. We then proceed to elaborate on the key components. 

\subsection{Multi-Source Information Exploration}
Our goal is to develop a model that can efficiently predict the labels of numerous unlabeled texts when trained on a given short text dataset $\mathcal{D}$ with limited labeled samples. To alleviate the issue of semantic sparsity in short texts, we aim to perform multi-source information exploration to maximally utilize statistical and linguistic information from the text itself, as well as factual information from outside.

\noindent \textbf{Statistical Information}: As mentioned before, statistical information is related to word statistics in the text. To capture this information, we construct a word graph $\mathcal{G}_w=\{\mathcal{V}_w, \mathbf{X}_w, \mathbf{A}_w\}$, where $\mathcal{V}_w$ is the set of word nodes and $\mathbf{A}_w \in \mathbb{R}^{|\mathcal{V}_w| \times |\mathcal{V}_w|}$ is the corresponding adjacency matrix determined by point-wise mutual information (PMI), \textit{i.e.}, $\mathbf{A}_{w,ij}=\max(\text{PMI}(v_i,v_j), 0)$, where $v_i, v_j \in \mathcal{V}_w$, which is a popular way to measure the word co-occurrence relationship. $\mathbf{X}_w \in \mathbb{R}^{|\mathcal{V}_w| \times f_w}$ is the feature matrix of all words with $f_w$-dimensional features. We initialize $\mathbf{X}_w$ as pretrained word embeddings generated by the GloVe method, which explicitly utilizes the global co-occurrence information of words and captures the underlying statistical information. Then, we feed the word graph $\mathcal{G}_w$ into the GCN using Eq.\ref{gnn} to obtain updated node embeddings $\mathbf{H}_w$ with statistical information.

\noindent \textbf{Linguistic Information}: This information is necessary for comprehensively understanding short texts, including semantic and syntactic structure. Here, we acquire linguistic information by identifying the syntactic roles of each word in a given text, such as adjectives or adverbs, which helps to eliminate syntactic word ambiguity. To this end, we construct a POS graph $\mathcal{G}_p=\{\mathcal{V}_p,\mathbf{X}_p, \mathbf{A}_p\}$, where $\mathcal{V}_p$ is the formed POS tag node set and $\mathbf{A}_p$ denotes the POS adjacency matrix calculated by PMI, \textit{i.e.}, $\mathbf{A}_{p,ij}=\max(\text{PMI}(v_i,v_j), 0)$, where $v_i, v_j \in \mathcal{V}_p$. We initialize the node features $\mathbf{X}_t \in \mathbb{R}^{|\mathcal{V}_t| \times f_t}$ as one-hot vectors. Similarly, we obtain updated POS tag features $\mathbf{H}_p$ by performing Eq.\ref{gnn}.

\noindent \textbf{Factual Information}: Additional factual information can help supplement the contextual knowledge required for short texts to enhance the classification ability of subsequent models. Therefore, we extract the entities in the short text that are resided in the knowledge graph and construct an entity graph $\mathcal{G}_e=\{\mathcal{V}_e,\mathbf{X}_e, \mathbf{A}_e\}$. Here, we utilize the TAGME tool for entity linking on the NELL \cite{andrew2010toward} knowledge graph. $\mathcal{V}_e$ is the entity node set. The entities' embeddings $\mathbf{X}_e \in \mathbb{R}^{|\mathcal{V}_e|\times f_e}$ are initialized by TransE \cite{bordes2013translating}. $\mathbf{A}_e$ is the entity adjacency matrix derived by the cosine similarity of each entity pair, \textit{i.e.}, $\mathbf{A}_{e,ij}={\max(\cos(\mathbf{X}_{e,i},\mathbf{X}_{e,j}),0)}$. The updated entity node embeddings $\mathbf{H}_e$ are also obtained by performing Eq.\ref{gnn}. 

\subsection{Text Representation Learning}
Given three types of graphs $\mathcal{G}=\{\mathcal{G}_\pi\}, {\pi \in \{w,e,p\}}$, to obtain text embeddings, we employ the following information aggregation strategy:
\begin{equation}
\begin{aligned}
\label{text}
    \mathbf{Z}_\pi &= \mathbf{P}_\pi\mathbf{H}_\pi, \\ 
    \mathbf{Z}_\pi &= \mathbf{Z}_\pi / ||\mathbf{Z}_\pi||_2, \pi \in \{w,e,p\},
\end{aligned}
\end{equation}
where $\mathbf{H}_\pi$ denotes the updated node embeddings of $\mathcal{G}_\pi$ obtained via a 2-layer GCN. We set $\mathbf{P}_\pi \in \mathbb{R}^{N \times |\mathcal{V}_\pi|}$ as the TF-IDF value between each text and word or POS tag of the corpus when $\pi \in \{w,p\}$. $N$ denotes the number of short texts. Moreover, when $\pi=e$, we make $\mathbf{P}_{e,ij}=1$ if the $i$-th text contains the $j$-th entity and 0 otherwise. After the normalized text-relevant features $\mathbf{Z}_w$, $\mathbf{Z}_e$, and $\mathbf{Z}_p$ are derived, we concatenate them to obtain the text embeddings, \textit{i.e.}, $\mathbf{Z}=\mathbf{Z}_w || \mathbf{Z}_e || \mathbf{Z}_p$. 

\subsection{Data Augmentation}
A key step for applying CL to NLP is to construct positive sample pairs. A typical approach for generating positive samples is a data augmentation technique, such as back-translation \cite{edunov2018understanding}, random noise injection \cite{xie2019data}, and word substitution \cite{wei2019eda}. Here, we augment the original data by replacing its words with WordNet synonyms. Formally, for each text $d_i^{\text{org}}$ in the original corpus $\displaystyle \mathcal{D}_{\text{org}} =\left\{d_{i}^{\text{org}}\right\}_{i=1}^{N}  $, we can obtain the augmented text $\displaystyle d_{i}^{\text{aug}} =\text{aug}\left( d_{i}^{\text{org}}\right)$ and augmented corpus $\displaystyle \mathcal{D}_{\text{aug}} =\left\{d{_{i}^{\text{aug}}}\right\}_{i=1}^{N}$. We denote the overall corpus and the corresponding text embeddings as $\displaystyle \mathcal{D} =\mathcal{D}_{\text{org}} \cup \mathcal{D}_{\text{aug}}$ and $\mathbf{Z}=\mathbf{Z}^{\text{org}}\cup\mathbf{Z}^{\text{aug}}$, 
respectively.
Note that our model is feasible for data augmentations, and we explore the impacts of different data augmentations on the model in the experiment section. 

\subsection{Hierarchical Structure among Tasks}
In contrast to prior models, we have implemented a \textit{hierarchical structure} to explicitly account for the relationship established through distinct stages of learned text features among the primary classification task and the auxiliary CL tasks. First, we utilize the rudimentary features acquired during the multi-source information exploration stage to perform ICL, enabling us to capture fine-grained contrastive information. Then, based on the intermediate features obtained at the ICL stage, we perform CCL to capture coarse-grained contrastive information. Finally, leveraging abstract features obtained at the CCL stage, we carry out the ultimate classification task. 

\noindent \textbf{Instance-Level Contrastive Learning}: First, based on the rudimentary text features $\mathbf{Z}$, we leverage ICL to perform instance discrimination tasks to explore fine-grained contrastive information. Typically, two texts from the same source data exhibit similar meanings. Their encoded text-level embeddings should be as similar as possible in the latent space. 
We refer to $d_i$ and the augmented version $d_j$ as a pair of positive samples while treating the other $2(N-1)$ texts in $\mathcal{D}$ as negative samples to this positive pair, which should be far away from the positive samples. With the obtained text embeddings $\mathbf{Z}$, we perform the normalization operation on them, 
\textit{i.e.}, $\tilde{\mathbf{Z}}=\mathbf{Z}/||\mathbf{Z}||_2$. Notably, we do not map $\mathbf{Z}$ to a hidden space through a projection head as in traditional CL. Due to the fine-grained information required by ICL, introducing a projection head would not only compromise the semantics but also introduce more parameters. 
Therefore, we avoid using a projection head in this stage. The objective function for a positive pair of examples ($d_i, d_j$) is defined as follows:
\begin{equation}
\label{icl}
    \mathcal{L}^{\text{ICL}}_i = -\log\frac{\exp((\tilde{\mathbf{Z}}_i\cdot \tilde{\mathbf{Z}}_j)/\tau)}{\sum_{k=1}^{2N} \mathbb{I}_{k\neq i}\exp((\tilde{\mathbf{Z}}_i \cdot \tilde{\mathbf{Z}}_k)/\tau)},
\end{equation}
where $\tilde{\mathbf{Z}}_i$ and $\tilde{\mathbf{Z}}_j$ denote the output embeddings of the $i$-th text and its augmented text, respectively. $\mathbb{I}_{k\neq i}$ is an indicator function set to 1 if $k\neq i$, and $\tau$ denotes the temperature parameter.

The ICL loss is computed by averaging over all positive pairs on $\mathcal{D}$, which is expressed as:
\begin{equation}
\label{totalicl}
    \mathcal{L}^{\text{ICL}}=\frac{1}{2N}\sum\nolimits_{i=1}^{2N}\mathcal{L}_i^{\text{ICL}}.
\end{equation}

\noindent \textbf{Cluster-Level Contrastive Learning}:
Next, based on the intermediate text features $\tilde{\mathbf{Z}}$ derived in the ICL stage corresponding to the corpus $\mathcal{D}$, we perform CCL. We expect to assign pseudo-cluster labels to the data in the corpus to explore their similarities from a cluster perspective and exploit their high-level feature clustering information such that similar instances can be pulled together. 
For ease of the presentation, we denote $\tilde{\mathbf{Z}}^\ast$ as the original or augmented text features from $\mathcal{D}_\ast$, where $\ast$ stands for ``org'' or ``aug''. We leverage the scores computed by the cosine similarity function to build relations between different texts. Next, we define a text $d_j^\ast$ as the nearest neighbor of text $d_i^\ast$ when Eq.\ref{near} is satisfied.
\begin{equation}
\label{near}
\begin{aligned}
    \text{near}(d_i^\ast)=\arg\max\nolimits_{d_j^\ast}\cos(\tilde{\mathbf{Z}}_i^\ast,\tilde{\mathbf{Z}}_j^\ast) \\ 
    \textit{s.t.} \quad \forall d_j^\ast\in\mathcal{D}_\ast\land{j\neq i}.
\end{aligned}
\end{equation}
 The text $d_i^\ast$ is connected with the text $d_j^\ast$ if $\text{near}(d_i^\ast)=d_j^\ast$ or $\text{near}(d_j^\ast)=d_i^\ast$. After performing the above operation, we can acquire the symmetric connections within $\mathcal{D}_\ast$. Then, we utilize the connected component labeling algorithm \cite{di1999simple} to assign the corresponding pseudo-cluster label for each derived component. Since any two instances in a component can be connected by paths, we treat its internal instances as similar. Subsequently, we can obtain the label matrix $\mathbf{Y}^\ast$. $\mathbf{Y}^\ast_{ij}$ is set to 1 if $d_i^\ast$ and $d_j^\ast$ are in the same component.
 
 Subsequently, we adopt a projection head $\Phi(\cdot)$, which maps the representations $\tilde{\mathbf{Z}}$ to a hidden space where the CCL loss is applied, \textit{i.e.}, $\mathbf{U}=\Phi(\tilde{\mathbf{Z}})$. Due to the different target granularities, there may be potential conflicts in the feature space between CCL and ICL, thus requiring a projection head. Here, 
 the dimension of $\mathbf{U}$ is half the dimension of $\tilde{\mathbf{Z}}$. Next, we normalize the output into a unit form, \textit{i.e.}, $\tilde{\mathbf{U}}=\mathbf{U}/||\mathbf{U}||_2$. Moreover, intuitively, since $d_i^{\text{org}}$ and $d_i^{\text{aug}}$ have the same meanings, the labels $\mathbf{Y}_i^{\text{org}}$ and $\mathbf{Y}_i^{\text{aug}}$ should be consistent. We can swap supervision signals between them. Another crucial reason for swapping supervision is that the dot product value of the positive samples in the same class may be large, which leads to a potentially small $\mathcal{L}_i^{\text{CCL}}$ under standard CL settings, thus affecting model optimization. The detailed CCL loss with swapped supervision is defined as follows:
\begin{equation}
\label{ccl}
\small
\begin{aligned}
    \mathcal{L}^{\text{CCL}}&=\frac{1}{N}\sum\nolimits_{i=1}^N\mathcal{L}^{\text{CCL}}_i, \\
    \mathcal{L}^{\text{CCL}}_i&=\mathcal{L}_i^{\text{swap}}(\mathbf{U}_{\text{org},i}, \mathbf{Y}^{\text{aug}}_i) + \mathcal{L}_i^{\text{swap}}(\mathbf{U}_{\text{aug},i}, \mathbf{Y}^{\text{org}}_i), \\
    \mathcal{L}_i^{\text{swap}}&=-\sum_j^N\mathbb{I}_{\mathbf{Y}_{ij}=1}\log\frac{\exp(\mathbf{U}_i \cdot \mathbf{U}_j/\tau)}{\sum\nolimits_{k=1}^N\mathbb{I}_{k\neq i}\exp(\mathbf{U}_i \cdot \mathbf{U}_k/\tau)},
\end{aligned}
\end{equation}
where $\tau$ symbolizes the temperature parameter and ``$\cdot$'' denotes the dot product operator. $\mathbb{I}_{\mathbf{Y}_{ij}=1}$ aims to find texts with the same label as that of the $i$-th text.

In these two applied components, ICL can provide beneficial information for the subsequent CCL task, while CCL can offer further guidance for the final classification task. Moreover, they can form a complementary relationship: ICL can provide a certain degree of variance in the obtained features, which can prevent the feature variability collapse. CCL has the capacity to mitigate class collision to a certain extent. 

\noindent \textbf{Classification Task}: Finally, leveraging the abstract text features $\tilde{\mathbf{U}}$ derived from the CCL process, we perform the final classification task. 
We adopt an extra projection head $\Psi(\cdot)$ with the same structure as that of $\Phi(\cdot)$ to map the text embeddings $\tilde{\mathbf{U}}$ to another latent space, \textit{i.e.}, $\mathbf{Q}=\Psi(\tilde{\mathbf{U}})$. Here, the dimension of $\mathbf{Q}$ is the number of classes. Then, we make predictions of these labeled data by performing a linear transformation followed by a ReLU activation on their hidden features. We specify the loss function as the commonly used cross-entropy function. Formally, the above operations can be expressed as:
\begin{equation}
\label{ce}
\begin{aligned}
    \mathcal{L}^{\text{CE}}&=-\frac{1}{\vert\mathcal{D}_{\text{train}}\vert}\sum_{i\in \mathcal{D}_{\text{train}}}\sum_j^c\mathcal{Y}_{ij}\log\mathbf{Q}_{ij}
\end{aligned}
\end{equation}
where $\mathcal{D}_{\text{train}}$ is the set of training data from $\mathcal{D}_{\text{org}}$ and $\mathcal{Y}$ is the one-hot vector of the ground-truth label of the training data. 
$c$ is the number of classes.

The adopted hierarchical architecture of tasks is similar to multi-task learning \cite{zhang2021survey}, which offers several advantages. First, by gradually progressing from ICL to CCL and then to classification, the model can extract more abstract and high-level features, which helps improve its generalization ability. Second, through step-by-step learning, the model can better utilize the data, as the results of the previous stage can provide better initialization and guidance for the subsequent stage. This inter-task correlation is beneficial for the model learning.

\subsection{Model Optimation}

Overall, the final loss function of our proposed model is the combination of the classification loss $\mathcal{L}^{\text{CE}}$, ICL loss $\mathcal{L}^{\text{ICL}}$, and CCL loss $\mathcal{L}^{\text{CCL}}$, which is formulated as follows:
\begin{equation}
\label{finalloss}
    \mathcal{L} = \mathcal{L}^{\text{CE}} + \eta\mathcal{L}^{\text{ICL}} + \zeta\mathcal{L}^{\text{CCL}},
\end{equation}
where $\eta$ and $\zeta$ are hyperparameters that control the proportions of different losses.

During the model inference, we feed the obtained test text embeddings to the classification head $\Upsilon(\cdot)$ to evaluate the model performance. 


\section{Experiment}
\noindent \textbf{Datasets}: We perform experiments on real-world STC datasets employed in earlier studies \cite{linmei2019heterogeneous,wang2021hierarchical}, \textit{i.e.}, \textbf{Twitter}, \textbf{MR} \cite{pang2005seeing}, \textbf{Snippets} \cite{phan2008learning}, \textbf{Ohsumed} \cite{hersh1994ohsumed}, and \textbf{TagMyNews} \cite{vitale2012classification}. The statistics of these datasets are summarized in Table \ref{data}. 

\begin{table*}[ht]
\centering
\begin{tabular}{c|c|c|c|c|c|c|c}
\hline
Dataset   & \#Docs  & \#Train (ratio) & \#Words & \#Entities & \#Tags & Avg.Len & \#Classes \\ \hline
Twitter   & 10,000 & 40 (0.40\%)    & 21,065 & 5,837    & 41    & 3.5        & 2       \\
MR        & 10,662 & 40 (0.38\%)    & 18,764 & 6,415    & 41    & 7.6        & 2       \\
Snippets  & 12,340 & 160 (1.30\%)   & 29,040 & 9,737    & 34    & 14.5       & 8       \\
Ohsumed   & 7,400  & 460 (6.22\%)   & 11,764 & 4,507    & 38    & 6.8        & 23      \\
TagMyNews & 32,549 & 140 (0.43\%)   & 38,629 & 14,734   & 42    & 5.1        & 7       \\ \hline
\end{tabular}
\caption{Summary statistics of the evaluation datasets.}
\label{data}
\end{table*}


The preprocessing for these datasets is consistent with previous studies, including the removal of non-English characters, stop words, and infrequent words with counts of less than five. Following previous studies \cite{linmei2019heterogeneous}, we randomly sample 40 labeled short documents per class, half of which form the training set, and the other half form the validation set. The remaining data constitute the test set, and their labels are invisible during training.

\noindent \textbf{Baselines}: We select four types of baseline models for comparison. (1) Traditional models include \textbf{TF-IDF+SVM} and \textbf{PTE} \cite{tang2015pte}. (2) Deep learning models contain \textbf{CNN}s \cite{kim-2014-convolutional}, \textbf{LSTM} \cite{liu2015multi} and \textbf{BERT} \cite{devlin2019bert}. Here, \textbf{BERT-avg} and \textbf{BERT-cls} denote the text embeddings represented by the average word embeddings and the token CLS embedding, respectively. (3) GNN-based models consist of \textbf{TLGNN} \cite{huang2019text}, \textbf{HyperGAT} \cite{ding2020more}, \textbf{TextING} \cite{zhang2020every}, \textbf{DADGNN} \cite{liu2021deep}, and \textbf{TextGCN} \cite{yao2019graph}. (4) Deep short text models include \textbf{STCKA} \cite{chen2019deep}, \textbf{HGAT} \cite{linmei2019heterogeneous}, \textbf{STGCN} \cite{ye2020document}, \textbf{SHINE} \cite{wang2021hierarchical}, \textbf{NC-HGAT} \cite{go2022contrastive}, and \textbf{GIFT} \cite{liu2024improved}. Notably, we also provide several large language models, containing \textbf{GPT-3.5} \cite{ouyang2022training}, \textbf{Bloom-7.1B} \cite{workshop2022bloom}, \textbf{Llama2-7B} \cite{touvron2023llama}, and \textbf{Llama3-8B} \cite{llama3modelcard}. Due to computational resource constraints, we only fine-tune approximately 7B LLMs through some GPU reduction techniques. 

\noindent \textbf{Evaluation Metric}: We use the accuracy (ACC) and macro-F1 score (F1) to evaluate the model performance, which are widely adopted by previous studies \cite{linmei2019heterogeneous}. All experiments are repeated ten times to obtain average metrics.

\section{Result}
\begin{table*}[ht]
\centering
\begin{tabular}{@{}c|cc|cc|cc|cc|cc@{}}
\toprule
\multirow{2}{*}{Model} & \multicolumn{2}{c|}{Twitter}    & \multicolumn{2}{c|}{MR}         & \multicolumn{2}{c|}{Snippets}   & \multicolumn{2}{c|}{Ohsumed} & \multicolumn{2}{c}{TagMyNews}   \\ \cmidrule(l){2-11} 
                       & ACC            & F1             & ACC            & F1             & ACC            & F1             & ACC            & F1          & ACC            & F1             \\ \midrule
TF-IDF+SVM             & 53.62          & 52.46          & 54.29          & 48.13          & 64.70          & 59.17          & 39.02          & 24.78       & 39.91          & 32.05          \\
PTE                    & 54.24          & 53.17          & 55.02          & 52.62          & 63.10          & 59.11          & 38.29          & 22.27       & 40.39          & 34.12          \\ \midrule
CNN                    & 57.29          & 56.02          & 59.06          & 59.01          & 77.09          & 69.28          & 32.92          & 12.06       & 57.12          & 45.37          \\
LSTM                   & 60.28          & 60.22          & 60.89          & 60.70          & 75.89          & 67.72          & 28.86          & 7.20        & 57.32          & 45.56          \\
BERT-avg               & 54.92          & 51.16          & 51.69          & 50.65          & 79.31          & 78.47          & 24.29          & 5.65        & 55.11          & 44.31          \\
BERT-cls               & 52.06          & 43.41          & 53.50          & 47.02          & 81.55          & 79.06          & 22.26          & 5.50        & 58.19          & 42.35          \\ \midrule
TLGNN                  & 59.02          & 54.56          & 59.22          & 59.36          & 70.25          & 63.29          & 35.76          & 13.12       & 45.25          & 33.52          \\
HyperGAT               & 59.15          & 55.19          & 58.65          & 58.62          & 70.89          & 63.42          & 36.60          & 20.02       & 45.60          & 31.51          \\
TextING                & 59.62          & 59.22          & 58.89          & 58.76          & 71.10          & 70.65          & 38.26          & 21.35       & 52.10          & 39.99          \\
DADGNN                 & 59.51          & 55.32          & 58.92          & 58.86          & 71.65          & 70.66          & 37.65          & 22.16       & 47.96          & 39.25          \\
TextGCN                & 60.15          & 59.82          & 59.12          & 58.98          & 77.82          & 71.95          & 41.56          & 27.43       & 54.28          & 46.01          \\ \midrule
STCKA                  & 57.56          & 57.02          & 53.25          & 51.19          & 68.96          & 61.27          & 32.20          & 12.25       & 32.15          & 23.26          \\
HGAT                   & 63.21          & 62.48          & 62.75          & 62.36          & 82.36          & 74.44          & 42.68          & 24.82       & 61.72          & 53.81          \\
STGCN                  & 64.33          & 64.29          & 58.25          & 58.22          & 70.01          & 69.93          & 35.22          & 28.30       & 35.65          & 35.16          \\
SHINE                  & 72.54          & 72.19          & 64.58          & 63.89          & 82.39    & 81.62    & 45.57          & 30.98       & 62.50          & 56.21          \\
NC-HGAT                  & 63.76          & 62.94          & 62.46          & 62.14          & 82.42    & 74.62    & 43.27          & 27.98       & 62.15          & 55.02 \\
GIFT                    & \underline{73.16} & \underline{73.16} & \underline{65.21} & \underline{65.21} & \underline{83.73} & \underline{82.35} & \underline{45.62} & \underline{31.25} & \underline{63.26} & \underline{56.92} \\ \midrule
Ours                   & \textbf{75.11}          & \textbf{75.06}          & \textbf{66.49}          & \textbf{66.47}          & \textbf{87.90} & \textbf{86.84} & \textbf{48.56} & \textbf{33.20} & \textbf{69.72} & \textbf{65.94} \\ \midrule
\textcolor{gray}{GPT-3.5}                & \textcolor{gray}{81.23}          & \textcolor{gray}{80.02}          & \textcolor{gray}{87.43} & \textcolor{gray}{86.62}          & \textcolor{gray}{66.52}          & \textcolor{gray}{63.48}          & \textcolor{gray}{47.98} & \textcolor{gray}{32.49} & \textcolor{gray}{61.43}          & \textcolor{gray}{54.79}          \\
\textcolor{gray}{Bloom-7.1B}               & \textcolor{gray}{87.52} & \textcolor{gray}{86.56} & \textcolor{gray}{87.03}          & \textcolor{gray}{86.96} & \textcolor{gray}{71.39}          & \textcolor{gray}{60.76}          & \textcolor{gray}{37.46}          & \textcolor{gray}{30.12}       & \textcolor{gray}{66.13}          & \textcolor{gray}{62.13}          \\
\textcolor{gray}{Llama2-7B}             & \textcolor{gray}{87.45}    & \textcolor{gray}{86.43}    & \textcolor{gray}{87.26}    & \textcolor{gray}{86.69}    & \textcolor{gray}{73.05}          & \textcolor{gray}{68.11}          & \textcolor{gray}{42.16}          & \textcolor{gray}{30.19}       & \textcolor{gray}{67.31}    & \textcolor{gray}{64.33}    
 \\ 
\textcolor{gray}{Llama3-8B}             & \textcolor{gray}{89.50}    & \textcolor{gray}{89.47}    & \textcolor{gray}{84.38}    & \textcolor{gray}{84.16}    & \textcolor{gray}{61.68}          & \textcolor{gray}{60.62}          & \textcolor{gray}{38.76}          & \textcolor{gray}{22.93}       & \textcolor{gray}{65.80}    & \textcolor{gray}{60.69}    
 \\ \bottomrule
\end{tabular}%
\caption{Results (\%) of several the Accuracy and Macro-F1 score on several short text datasets. We highlight the best performance in bold excluding the LLMs based on the pairwise t-test with 95\% confidence.}
\label{result}
\end{table*}

\noindent \textbf{Model Performance}: Table \ref{result} indicates that MI-DELIGHT achieves competitive performance across several datasets by a large margin in terms of accuracy and macro-F1 score. 
A key factor contributing to the remarkable superiority of MI-DELIGHT over other competing models lies in its deliberate design of a dual CL auxiliary task. This task serves the purpose of acquiring informative text representations and effectively capturing contrastive information at various levels. Specifically, the ICL and CCL within this framework enable the model to discern fine-grained details while also considering broader patterns and contexts. 
Moreover, the introduced hierarchical concept can help the model learn step by step, while fully utilizing the inter-task correlations, thereby enhancing the overall model performance. Moreover, we construct three types of graphs, including a word graph, a POS graph, and an entity graph, to incorporate statistical, linguistic and factual knowledge, which exploits semantic and syntactic information from the text and additional information from outside. All of the above operations are beneficial for better identifying the correct meanings of short texts.

We find that MI-DELIGHT basically achieves greater improvements on the Snippets, Ohsumed, and TagMyNews, which even considerably surpasses LLMs on these datasets. 
We attribute this phenomenon to the fact that the more unlabeled texts exist, the better MI-DELIGHT can extract useful self-supervised signals from them. LLMs have limited understanding in specific-domain (\textit{e.g.}, medical domain) texts. However, LLMs achieve satisfactory performance in general texts such as Twitter and MR. This is because they are pre-trained on vast amounts of high-quality data and have a large number of parameters.  
Moreover, they may have already encountered part of the test data.


\begin{table*}[!ht]
\centering
\begin{tabular}{@{}c|cc|cc|cc|cc|cc@{}}
\toprule
\multirow{2}{*}{Model} & \multicolumn{2}{c|}{Twitter} & \multicolumn{2}{c|}{MR} & \multicolumn{2}{c|}{Snippets} & \multicolumn{2}{c|}{Ohsumed} & \multicolumn{2}{c}{TagMyNews} \\ \cmidrule(l){2-11} 
                       & ACC           & F1           & ACC        & F1         & ACC           & F1            & ACC           & F1           & ACC           & F1            \\ \midrule
\textit{w/o word graph} & 62.60 & 62.10 & 55.02 & 54.96 & 76.15 & 74.80 & 29.08 & 21.26 & 62.52 & 57.19 \\
\textit{w/o POS graph} & 72.40 & 71.65 & 65.46 & 65.25 & 85.86 & 85.19 & 46.96 & 28.19 & 66.39 & 59.10 \\
\textit{w/o entity graph} & 70.42 & 70.36 & 64.76 & 64.79 & 85.95 & 85.12 & 47.55 & 29.98 & 67.28 & 62.90 \\
\textit{w/o CCL and ICL} & 72.57 & 72.26 & 62.33 & 62.32 & 86.12 & 83.89 & 45.98 & 27.53 & 66.38 & 62.10 \\
\textit{w/o CCL}        & 72.19         & 72.18        & 66.02      & 65.92      & 85.34         & 83.99         & 48.38         & 31.59        & 66.56         & 62.45         \\
\textit{w/o ICL}        & 73.91         & 73.60        & 65.43      & 65.29      & 85.79         & 83.48         & 46.29         & 28.96        &  66.41        & 62.34         \\
\textit{parallel}        & 73.74         & 73.72        & 65.32      & 65.30      & 84.54         & 83.82         & 48.51         & 31.82        &  68.72        & 64.94         \\
MI-DELIGHT (deletion)      & 72.04         & 71.92        & 63.52      & 63.50      & 83.98         & 82.26         & 44.59         & 27.92        &   67.44       &   62.67       \\
MI-DELIGHT (context)      & \textbf{75.66}         & \textbf{75.56}        & 65.68      & 65.62      & 84.01         & 82.66         & 44.29         & 28.25        & 67.72         & 63.28         \\
MI-DELIGHT (WordNet)      & 75.11         & 75.06        & \textbf{66.49}      & \textbf{66.47}      & \textbf{87.90}         & \textbf{86.84}         & \textbf{48.56}         & \textbf{32.20}        & \textbf{69.72}         & \textbf{65.94}         \\ \bottomrule
\end{tabular}%
\caption{The ablation and different text augmentation results (\%) of various experimental settings.}
\label{ablation}
\end{table*}
\noindent \textbf{Model Variants}:
\label{variant} 
To assess the effectiveness of each part of MI-DELIGHT, we design the following model variants to perform ablation experiments. (1) \textit{w/o word graph}: We remove the word graph that introduces statistical knowledge. (2) \textit{w/o POS graph}: We exclude the POS graph that incorporates linguistic knowledge. (3) \textit{w/o entity graph}: We delete the entity graph that contains factual knowledge. (4) \textit{w/o CCL and ICL}: We remove CCL and ICL simultaneously, leaving the text representation learning module which is combined with the cross-entropy loss for optimization. (5) \textit{w/o CCL}: We eliminate the CCL module to demonstrate the role of the ICL module. (6) \textit{w/o ICL}: We exclude the ICL module to confirm the efficiency of the CCL module. (7) \textit{parallel}: We simply add projection heads for all tasks and perform them in parallel. We obtain several findings by observing the results presented in the first seven rows shown in Table \ref{ablation}. First, when we delete any part of the model, the performance of MI-DELIGHT decreases significantly, illustrating that each part plays an essential role in our model. Second, three constructed graphs used to enrich the short text information bring different types of information that play an indispensable role. Since the word graph can provide the most fundamental semantic information, removing it would significantly reduce the model's performance. Third, both the CCL and ICL modules are designed to allow the model to learn more discriminative text representations. 
Finally, our proposed hierarchical architecture is superior to the parallel version, since it fully utilizes the inter-task correlations.

Moreover, we explore the impacts of different approaches for generating positive sample pairs in the model. (1) \textbf{MI-DELIGHT (deletion)}: It randomly deletes a fraction of the words in a given sentence to generate an enhanced positive sample. (2) \textbf{MI-DELIGHT (context)}: It leverages pretrained large-scale language models (\textit{e.g.}, BERT) to find a portion of the input text with suitable words for substitution. (3) \textbf{MI-DELIGHT (WordNet)}: It generates augmented positive pairs by replacing words of an input text with WordNet synonyms, which is the default experimental setting. The empirical results are shown in the last three rows of Table \ref{ablation}. As expected, the relevant metrics obtained across all the datasets drastically decrease when we adopt the deletion method for augmenting original texts. A plausible reason is deleting keywords from the original sentence changes its semantic information. 

\section{Conclusion}
In this work, we propose a novel model named MI-DELIGHT for STC. We build three types of graphs to introduce the statistical, linguistic, and factual information for enriching short texts. Also, we design a dual-level CL auxiliary tasks to capture multi-grained contrastive information. 
Moreover,  we leverage a hierarchical structure to capture inter-task correlations. 
The empirical results reveal that MI-DELIGHT consistently outperforms other baselines, including some popular LLMs, across several datasets.

\begin{links}
    \link{Code}{https://github.com/KEAML-JLU/MI-DELIGHT}
\end{links}

\section*{Acknowledgments}
This work is supported in part by funds from the National Key Research and Development Program of China (No. 2021YFF1201200), the National Natural Science Foundation of China (No. 62172187 and No. 62372209). Fausto Giunchiglia’s work is funded by European Union’s Horizon 2020 FET Proactive Project (No. 823783).

\bibliography{aaai25}

\end{document}